\documentclass[twoside,11pt]{article}
\usepackage{mlhc}
\usepackage{graphicx}
\usepackage{subcaption}
\usepackage{booktabs}
\usepackage{tikz}
\usepackage{enumitem}
\usepackage{amsmath}
\usepackage{gensymb}
\usepackage{bbding}
\usepackage[ruled,vlined]{algorithm2e}

\DeclareMathOperator*{\argmin}{arg\,min}
\usepackage{tikz}
\def\checkmark{\tikz\fill[scale=0.4](0,.35) -- (.25,0) -- (1,.7) -- (.25,.15) -- cycle;}

\makeatletter
\renewcommand*{\@fnsymbol}[1]{\ensuremath{\ifcase#1\or \,\dagger \or \ddagger\or
		\mathsection\or \mathparagraph\or \|\or **\or \dagger\dagger
		\or \ddagger\ddagger \else\@ctrerr\fi}}
\makeatother

\begin{document}

\title{Towards Vision-Based Smart Hospitals: A System for Tracking and Monitoring Hand Hygiene Compliance}

\author{\name Albert Haque\textsuperscript{1} \email ahaque@cs.stanford.edu \\
\name Michelle Guo\textsuperscript{1} \email mguo95@stanford.edu \\
\name Alexandre Alahi\textsuperscript{1,2} \email alexandre.alahi@epfl.ch \\
\name Serena Yeung\textsuperscript{1} \email syyeung@cs.stanford.edu \\
\name Zelun Luo\textsuperscript{1} \email zelunluo@stanford.edu \\
\name Alisha Rege\textsuperscript{1} \email amr6114@stanford.edu \\
\name Jeffrey Jopling\textsuperscript{3} \email jjopling@stanford.edu \\
\name Lance Downing\textsuperscript{3} \email ldowning@stanford.edu \\
\name William Beninati\textsuperscript{4} \email bill.beninati@imail.org \\
\name Amit Singh\textsuperscript{5} \email amsingh@stanfordchildrens.org \\
\name Terry Platchek\textsuperscript{5} \email tplatchek@stanfordchildrens.org \\
\name Arnold Milstein\textsuperscript{3} \email milstein@stanford.edu \\
\name Li Fei-Fei\textsuperscript{1} \email feifeili@cs.stanford.edu \\
\addr \textsuperscript{1}Department of Computer Science, Stanford University, United States \\
\addr \textsuperscript{2}Department of Civil Engineering, \'{E}cole Polytechnique F\'{e}d\'{e}rale de Lausanne, Switzerland \\
\addr \textsuperscript{3}Clinical Excellence Research Center, School of Medicine, Stanford University, United States \\
\addr \textsuperscript{4}TeleCritical Care, Intermountain Healthcare, United States \\
\addr \textsuperscript{5}Department of Pediatrics, School of Medicine, Stanford University, United States
} 
       
\maketitle
\renewcommand*{\thefootnote}{\fnsymbol{footnote}}
\renewcommand*{\thefootnote}{\arabic{footnote}}

\begin{abstract}
One in twenty-five patients admitted to a hospital will suffer from a hospital acquired infection.
If we can intelligently track healthcare staff, patients, and visitors, we can better understand the sources of such infections.
We envision a smart hospital capable of increasing operational efficiency and improving patient care with less spending.
In this paper, we propose a non-intrusive vision-based system for tracking people's activity in hospitals. We evaluate our method for the problem of measuring hand hygiene compliance.
Empirically, our method outperforms existing solutions such as proximity-based techniques and covert in-person observational studies.
We present intuitive, qualitative results that analyze human movement patterns and conduct spatial analytics which convey our method's interpretability. 
This work is a step towards a computer-vision based smart hospital and demonstrates promising results for reducing hospital acquired infections.
\end{abstract}

\section{Introduction}
In recent years, smart hospitals have garnered large research interest in the healthcare community \citep{ma2017measuring, twinanda2015data, chakraborty2013video, sanchez2008activity, weiser2010effect, fisher2008tracking}.
Smart hospitals encompass a variety of technology-based products for controlling, automating and optimizing workflows in the hospital \citep{sanchez2008activity, chakraborty2013video}. 
Surgery checklists are known to improve clinical outcomes \citep{weiser2010effect} and smart sensors can be used to automate safety procedures \citep{yu2012smart}.
More recently, computer vision has been used to automatically recognize clinical activities \citep{chakraborty2013video, ma2017measuring, twinanda2015data}.
With automatic and continuous monitoring, smart hospitals can locate inventory, identify patients, track healthcare workers, and increase operational efficiency \citep{fisher2008tracking, lenz2007support}, ultimately improving patient care quality \citep{gao2006vital}.

One significant problem that smart hospitals can help address is the prevention of hospital acquired infections (HAIs) \citep{cook2009assessing}.
One in twenty-five patients admitted to a hospital suffers from HAIs \citep{cdc2016}, costing hospitals billions of dollars per year \citep{zimlichman2013health}. Proper hand hygiene is an important part of preventing such HAIs, and smart hospitals that are constantly aware of the dynamic environment can track the movements of healthcare workers, monitor hand hygiene compliance, and provide feedback.

\begin{figure}[t]
    \centering
    \includegraphics[width=0.8\textwidth]{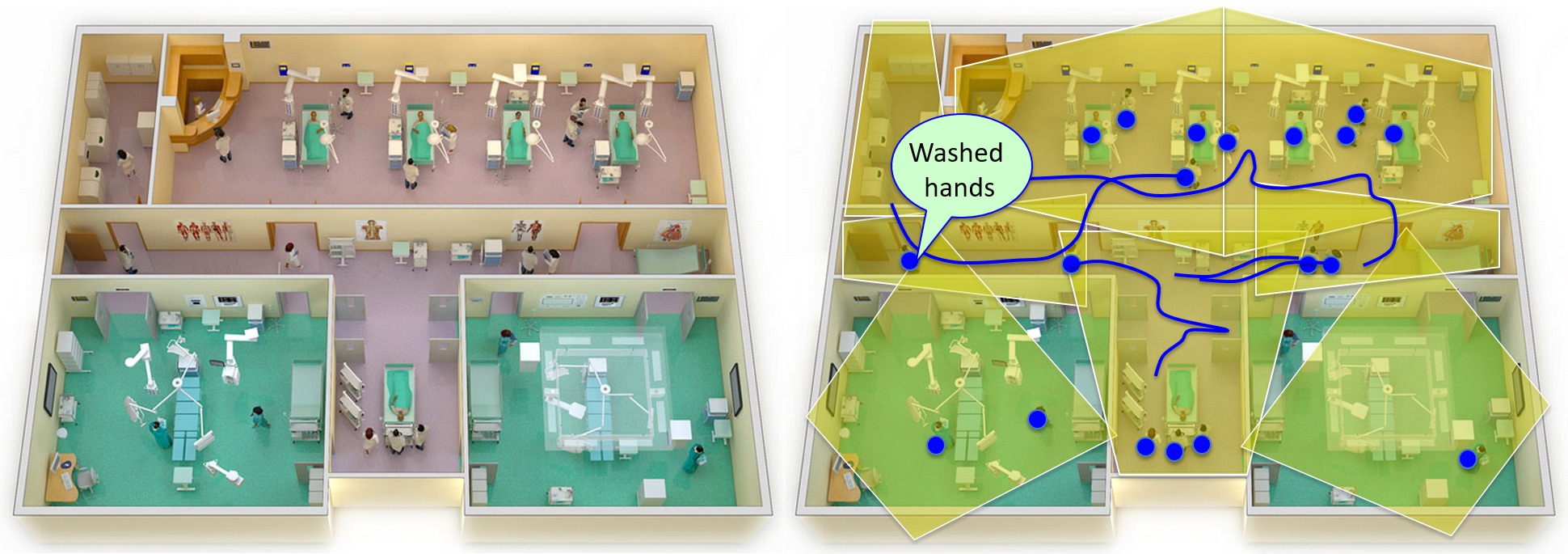}
    \caption{Illustration of our proposed vision-based smart hospital. (Left) Top-view of a hospital unit. (Right) Our system that track  fine-grained activities such as hand hygiene compliance. Yellow polygons indicate areas covered by our sensors. Blue dots indicate pedestrian detections and blue lines indicate pedestrian movements over time.}
    \label{fig:pull}
\end{figure}

Existing technologies for smart hospitals have been used to tackle some of these problems. Radio-frequency identification (RFID) systems have been used for tracking people in hospitals and are a common approach towards building a smart hospital \citep{fuhrer2006building, pineles2013rfid, polgreen2010method}. RFID systems are generally cheap and easy to deploy \citep{coustasse2013impact}.
Despite these benefits, objects and humans of interest must be manually tagged and remain physically close to a base station for the system to correctly locate them.
For people, this generally requires that hospital staff wear special wristbands, gloves, or external badges to register their position at specific events of interest (e.g., before entering a patient room) \citep{simmonds2011utility, yao2010use}.
While electronic counters attached to dispensers remove the badge requirement \citep{sahud2010electronic}, it still provides a non-continuous location history.
Combined with the noisy localization accuracy of indoor radio localization methods \citep{whitehouse2007practical, alahi2015rgbw, marra2014new, zanca2008experimental}, this makes RFID systems ill-suited for smart hospitals.
We need a more scalable solution to track both objects (e.g., portable computers, trash cans) and humans with higher precision at finer resolution (e.g., actions, body poses) and in a non-intrusive fashion.

We argue that computer vision-based approaches offer the ability to address these challenges. Recently, computer vision has pushed state-of-the-art in tracking applications such as self-driving cars \citep{cho2014multi} and sports analytics \citep{halvorsen2013bagadus, pers2000computer}.
They have the advantage of operating on a continuous data stream (i.e., videos) and have the ability to locate objects and pedestrians in both two- and three-dimensional space.
However, if we are to fully understand the hospital environment we must track more than just human locations.
Computer vision systems allow us to observe the world at high resolution, enabling the characterization of fine-grained motions such as handwashing or surgical procedures.
More recently, video cameras have been shown to reduce the Hawthorne effect, provide real-time feedback, and perform long-term monitoring in hospitals. \citep{armellino2013replicating, nishimura1999handwashing}.
However, video recording raises patient and staff privacy concerns, thereby limiting their use.
        
In this work, we develop a non-intrusive and privacy-safe vision-based system for tracking people’s activity in hospitals, specifically for the problem of hand hygiene.
While computer vision has shown promising results for tracking \citep{zhang2008global}, a number of unsolved challenges remain.
First, due to differences in building construction, hospitals may contain sparse networks of sensors with minimal overlapping field of views.
This can make it difficult for trajectory prediction and data association \citep{leibe2007coupled} across entire hospital units.
Second, hospitals exhibit variances in visitor and staff traffic patterns leading to extremely busy and crowded scenes.
Third, due to privacy laws surrounding personal health information, our system is limited in the visual data it can use.
We must resort to de-identified depth images, losing important visual appearance cues in the process.
This makes it especially difficult for our system to continuously locate and track personnel over time.
We must be able to link events of interest (e.g., exiting a room) with specific people.

In this paper, our contributions are as follows.
First, we propose a system for analyzing people's activities in hospitals.
We demonstrate our algorithm's performance on the task of automatic hand hygiene assessment in the context of a classification task and a tracking task.
Not only can this improve hand hygiene compliance  but it opens the doors for future smart hospital research.
Second, we present results analyzing physical movement patterns and hospital space analytics.
Our system's interpretable visualizations can be used to optimize nurse workflows, improve resource management \citep{fry2005mascal}, and monitor hygiene compliance.
Interpretability can help accelerate adoption time and build more trust between the computer, clinicians, and patients.

\section{Data}

Privacy regulations such as HIPAA and GDPR limit the use of cameras in healthcare settings.
In cases where cameras are allowed, access to recordings is strictly controlled, often preventing the use of computer vision algorithms.
To comply with these regulations, we use \textit{depth images} generated from depth sensors instead of standard color cameras. 

\begin{figure*}[t]
	\centering
	\begin{subfigure}[t]{0.25\textwidth}
		\includegraphics[width=0.99\textwidth]{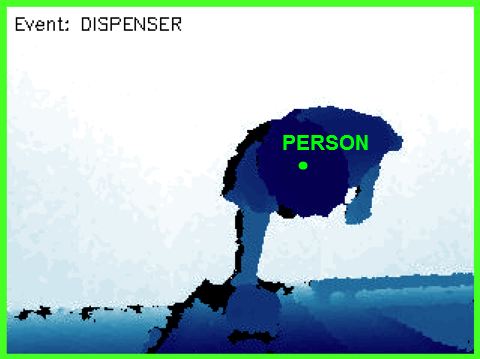}
	\end{subfigure}%
	\begin{subfigure}[t]{0.245\textwidth}
		\centering
		\includegraphics[width=1.0\textwidth]{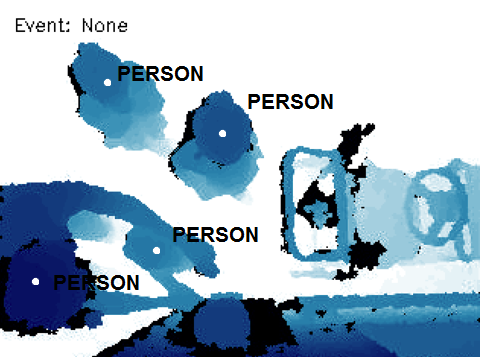}
	\end{subfigure}
	\begin{subfigure}[t]{0.245\textwidth}
		\centering
		\includegraphics[width=1.0\textwidth]{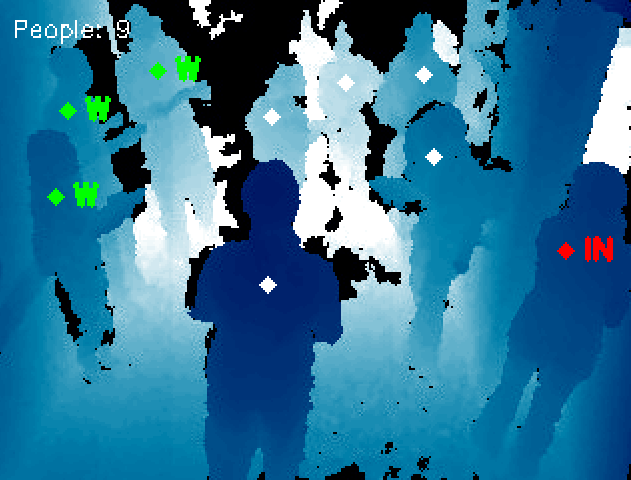}
	\end{subfigure}
	\begin{subfigure}[t]{0.245\textwidth}
		\centering
		\includegraphics[width=1.0\textwidth]{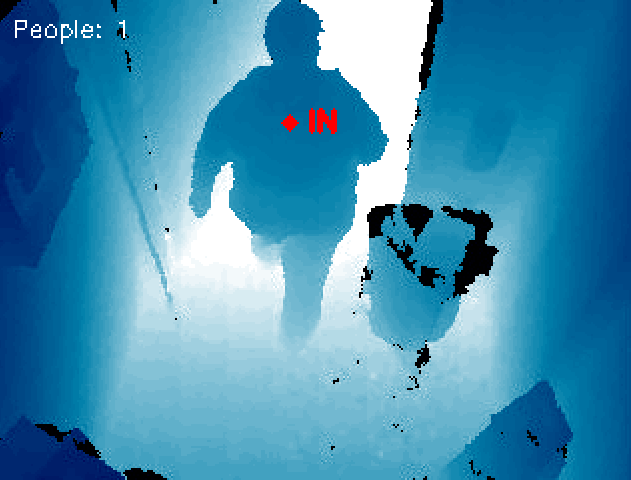}
	\end{subfigure}
	\caption{Example depth images with human annotations. White dots represent different people. Green dots or ``W" denote people performing handwashing events. Red ``IN" denotes a person entering a room. The green box denotes a binary classification label.}
	\label{fig:annotation}
\end{figure*}

\textbf{Depth Images.}
A \textit{depth image} can be interpreted as a de-identified version of a standard color photo (see Figure \ref{fig:annotation}).
In color photos, each pixel represents a color, often encoded as combination of red, green, and blue (RGB) values as unsigned integers.
In depth images, each pixel represents how far away the ``pixel" is, often encoded in real world meters as floating point numbers.
While we lose color information in depth images, we are able to respect patient privacy and data sharing protocols.
Notice that in depth images, we (humans) are unable to see the color of the person or the environment, we still understand the semantics of the scene (e.g., group of people standing in a hallway).
This is because depth images convey volumetric 3D information while color images convey colored appearance.
It is this 3D spatial information that allows our computer vision algorithm to reason about hand hygiene activity.

\textbf{Data Sources.} Images were collected from an acute care pediatric unit and an adult intensive care unit. A total of two hospitals participated in the data collection campaign.
Sensors were installed in top-down views above alcohol-based gel dispensers, in side-views overlooking the corridors, inside patient rooms, and in corners of anterooms overlooking sinks and patient beds.
The sensors have a field of view $58\degree$ horizontally, $45\degree$ vertically, and have a functional range between 0.8 and 4.0 meters. 
The images include objects such as garbage bins, computers, and medical equipment (see Figure \ref{fig:annotation}).

\textbf{Ground Truth.}
The ground truth, or \textit{gold standard}, is considered the absolute correct account or labeling as determined by one or more human subject matter experts, also known as \textit{annotators}.
Similar to \citet{armellino2012using}, the ground truth defines whether a person correctly followed hand hygiene protocol.
In hospitals, an annotator determines the ground truth either (1) on-site in real-time as events occur or (2) after-the-fact by watching video recordings.
A major benefit of annotating video recordings is the ability to move forward and backward in time.
Additionally, the annotator can replay the same event from multiple sensor viewpoints to verify correctness.
Because of these benefits, we chose to annotate our data using remote video recordings.

The ground truth was determined by ten annotators.
This group includes students, professors, and practicing physicians.
Each annotator was responsible for identifying the $(x,y)$ position of all people present in the image by clicking with a computer mouse.
Additionally, for each person present in the image, the annotator was asked to note whether the person was entering or exiting a patient room and if so, whether they washed their hands upon entry or exit (Figure \ref{fig:annotation}).
This was done with keyboard input (e.g., pressing 1 or 2).
Clinicians trained each annotator on proper hand hygiene protocol with live demonstrations.
Each depth image was annotated by one to three annotators and their ground truth assessments were cross-validated.
On average, each annotator spent three seconds annotating each image.

\textbf{Dataset Statistics.}
Ground truth hand hygiene compliance data was collected on a Friday from 12 pm to 1 pm.
During this hour, which is peak lunch time, there were a large number of visitors in the hospital.
A total of 351 ground truth tracks were collected and annotated, of which 170 involved a person entering a patient room, of which 30 were compliant (i.e., followed correct hand hygiene protocol). Of the 181 tracks exiting a room, 34 were compliant.
The data used for the classifier (Section \ref{sec:method_classifier}) consists of 150,400 images, of which 12,292 images contained pedestrians using the dispenser. The training set contained 80\% of the images with the remaining 20\% allocated to the test set.

\section{Method}\label{sec:method}

Our goal is to develop a system to automatically detect, track, and assess hand hygiene compliance.
The hope is that such a system can be used as part of a greater smart hospital ecosystem and ultimately prevent HAIs caused by improper hand hygiene.
To compute the hand hygiene compliance rate, we must perform three tasks: (1) detect the healthcare staff, (2) track them in the physical world, (3) and classify their hand hygiene behavior.
This will allow us to compute quantitative compliance metrics across the hospital unit.

\subsection{Pedestrian Detection}

The first step involves locating the 3D positions of the staff within the field-of-view of each sensor.
We use the same sparsity-driven formulation as in \cite{golbabaee2014scoop}.
An inverse problem is formulated to deduce pedestrian location points (i.e., an occupancy vector $x$) given a sparsity constraint on the ground occupancy grid.
Let $y$ be our observation vector (i.e., the binary foreground silhouettes), and $D$ the dictionary of atoms approximating the binary foreground silhouettes of a single person at all locations.
We aim to solve:
\begin{equation}\label{eq:human_detection}
x = \argmin_{x\in\{0,1\}} ||y - D x||_2^2 \text{\; s.t. \;} ||x||_0 < \epsilon_p,
\end{equation}
where $\epsilon_p$ is an upper bound of the sparsity level.
We use the Set Covering Object Occupancy Pursuit algorithm \citep{golbabaee2014scoop} to infer the occupancy map $x$.
At each iteration, the algorithm selects the atom of the dictionary which best matches the image.

\subsection{Tracking Across Cameras}\label{tracklet_generation}

Once pedestrians are located in the field-of-view of each camera, to track their hygiene status, we must track them across cameras.
Formally, we want to find the set of trajectories $X$, where each trajectory $x \in X$ is represented as an ordered set of detections, $L_x$, representing the detected coordinates of pedestrians.
Similarly, $L_x = (l_x^{(1)} ,...,l_x^{(n)} )$ is an ordered set of intermediate detections which are linked to form the trajectories.
These detections are ordered by the time of initiation.
The problem can be written as a maximum a-posteriori (MAP) estimation problem.
Next, we assume a Markov-chain model connecting every intermediate detection $l_x^{(i)}$ in the trajectory $X$, to the subsequent detection $l_x^{(i+1)}$ with a probability given by $P(l_x^{(i+1)} | l_x^{i})$.
We can now formulate the MAP task as a linear integer program by finding the flow $f$ that minimizes the cost $C$:
\begin{align}\label{eq:tracklet_generation}
\min_{f} & \quad C =  \sum_{x_i\in X}\alpha_i f_i + \sum_{x_i, x_j \in X}\beta_{ij} f_{ij} \qquad \text{s.t} \qquad f_i, f_{ij} \in (0,1) \text{\quad and \quad} f_{i} = \sum_{j}{f_{ij}},
\end{align}
where $f_i$ is the flow variable indicating whether the corresponding detection is a true positive, and $f_{ij}$ indicates if the corresponding detections are linked together.
The variable $\beta_{ij}$ denotes the transition cost given by $\log P (l_i|l_j)$ for the detection $l_i,l_j\in L$. The local cost $\alpha_i$ is the log-likelihood of an intermediate detection being a true positive.
In our case, we suppose that all detections have the same likelihood.
This is equivalent to the flow optimization problem, solvable in real-time with k-shortest paths  \citep{berclaz2011multiple}.

\subsection{Hand Hygiene Activity Classification}\label{sec:method_classifier}
\begin{figure}[t]
	\centering
	\includegraphics[width=0.95\textwidth]{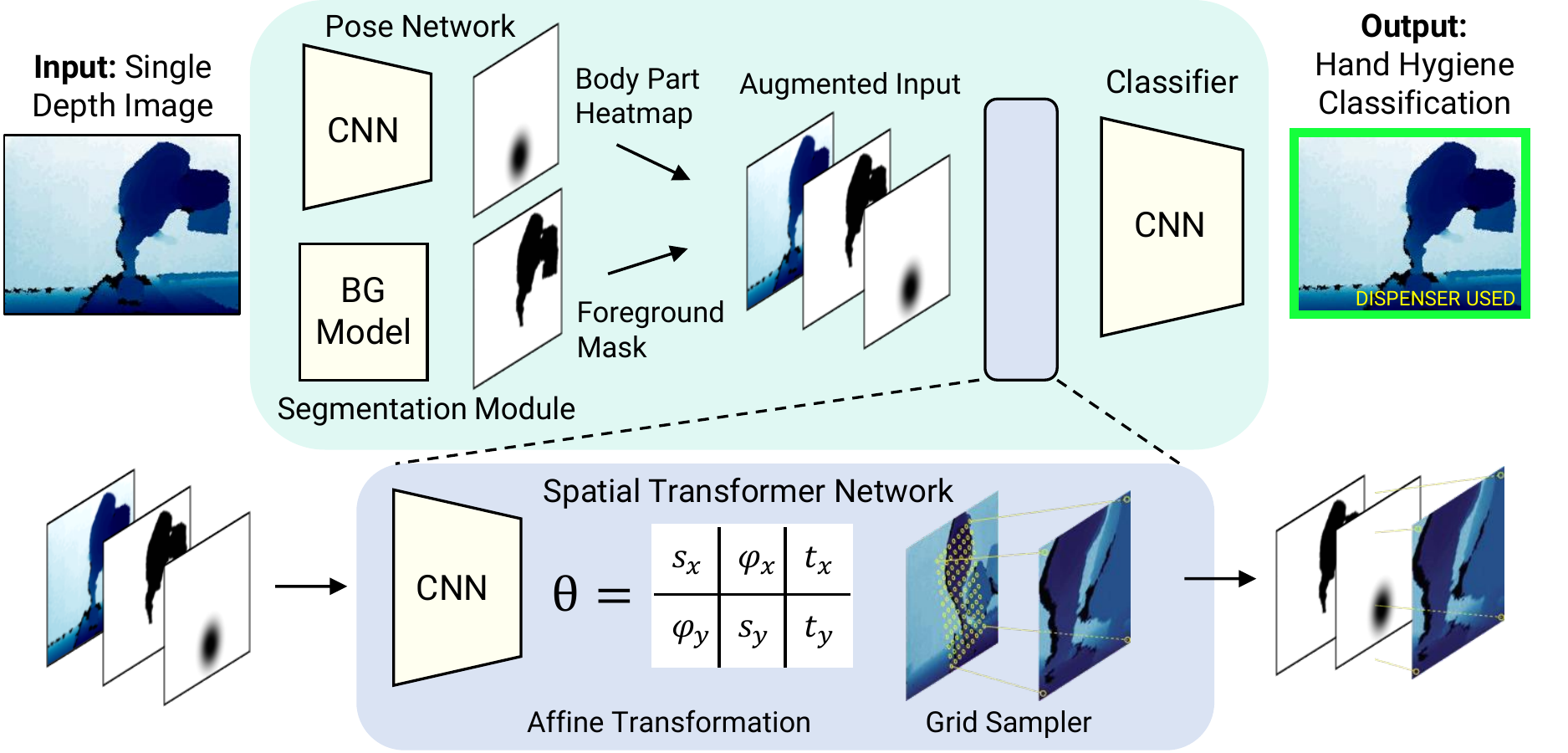}
	\caption{Hand hygiene activity classifier. Using pose and segmentation as mid-level cues, a spatial transformer projects the features into a viewpoint invariant feature space. $\theta$ denotes the learned transformation parameters: scale $(s_x, s_y)$, skew $(\varphi_x, \varphi_y)$, and translation $(t_x, t_y)$.}
	\label{fig:model}
\end{figure}

At this stage, we have identified the tracks (i.e., position on the global ground plane) of all pedestrians in a hospital unit.
The last step is to detect hand hygiene activity and link it to a specific track. That is, we must label each pedestrian track as \textit{clean} or \textit{not clean}.
A person becomes \textit{clean} when they use one of the various alcohol-based gel dispensers located throughout the hospital unit.
We propose an activity classifier which accepts depth images as input and outputs a binary prediction of whether a person used the dispenser or not. 

\textbf{Learned Viewpoint Invariance.}
Deployment of sensors in real-world settings is often prone to error.
Whether intentional or not, construction crews and maintenance technicians install sensors that vary in both angle and position.
If our goal is to propose a non-intrusive vision-based system for tracking hand hygiene activity, our model must be robust to such variances.
Towards this goal, the input to our classifier is a single depth image captured from any sensor viewpoint (Figure \ref{fig:model}).
We augment this depth image by estimating hand position estimates and a extracting a foreground body silhouette.
The pose network is a two-layer convolutional network and the segmentation module uses optical flow to determine foreground-background changes.
The output of the hand hygiene activity classifier is a binary prediction: whether a hand dispenser event occurred or not.

Traditional convolutional network architectures \citep{krizhevsky2012imagenet} are generally not viewpoint invariant \citep{jaderberg2015spatial}.
We address this issue by introducing an implicit attention mechanism: a fully differentiable spatial transformer network that can spatially transform any input feature map \citep{jaderberg2015spatial}.
Given the augmented input $U\in\mathbb{R}^{H\times W\times 3}$, the localization network $f(\cdot)$ predicts the two-dimensional transformation parameters $\theta$ needed to transform the input. This performs a geometric transformation on $U$ using the affine transformation matrix $A_\theta$ (also see Figure \ref{fig:model}):
\begin{equation}\label{eq:affine}
A_\theta = \begin{bmatrix}
\theta_1   &   \theta_2   &   \theta_3\\
\theta_4   &   \theta_5   &   \theta_6
\end{bmatrix}
=
\begin{bmatrix}
s_x   &   \varphi_x   &   t_x\\
\varphi_y   &   s_y   &   t_y
\end{bmatrix}
.
\end{equation}
With the transformation parameters $\theta$ predicted by the localization network, the grid generator computes a \textit{sampling grid} $G\in\mathbb{R}^{H\times W\times 3}$.
This grid identifies the points in the augmented input which will be used for the transformed output.
To perform a warping on $U$, a sampler applies a sampling kernel at each grid point $G_i=(x_i^t, y_i^t)$ to produce $V$, where $(x_i^t, y_i^t)$ are the target coordinates in $V$. We use a bilinear sampling kernel, where the kernel $k =$ max$(0, 1 - |\cdot|)$. The output feature map $V$ is thus defined as
\begin{equation}\label{eq:stn}
V_i^c = \sum_{n}^{H} \sum_{m}^{W} U_{n m}^{c} \text{ max}(0, 1 - |x_i^s - m|) \text{ max}(0, 1 - |y_i^s - n|)
\end{equation}
where $(x_i^s, y_i^s)$ are source coordinates in the augmented input $U$ that define the sample points.
Because Equation \ref{eq:stn} is a linear function applied to $U$, the spatial transformation network is differentiable, allowing loss gradients to flow through network \citep{jaderberg2015spatial}. At this point we have generated the transformed input $V$, which we will pass into a convolutional network classifier for predicting hand hygiene dispenser events.

\subsection{Fusing Tracking with Hand Hygiene Activity Classification}

Our goal is to create a system with a holistic understanding of the hospital.
In isolation, an activity detector is insufficient to understand human behavior over time and a tracker is unable to understand fine-grained motions on its own.
It is necessary to fuse the outputs from these components.
This becomes a spatio-temporal matching problem where we must match hand hygiene activity detections with specific tracks.

For each hand hygiene classifier detection (i.e., dispenser is being used), we must match it to a single track.
A match occurs between the classifier and tracker when a track $\mathcal{T}$ satisfies two conditions:
\begin{enumerate}[topsep=0pt,itemsep=0ex,partopsep=0ex,parsep=0ex]
    \item Track $\mathcal{T}$ contains $(x,y)$ points $\mathcal{P}$ which occur at the same time as the hand hygiene detection event $\mathcal{E}$, within some temporal tolerance level.
    \item At least one point $p \in \mathcal{P}$ is physically nearby the sensor responsible for the detection event $\mathcal{E}$. This is defined by a proximity threshold around the patient's door.
\end{enumerate}
If there are multiple tracks that satisfy these requirements, we break ties by selecting the track with the closest $(x, y)$ position to the door.

\textbf{Final Output.} The final output of our model is a list $T$ of tracks, where each track consists of an ordered list of $(t, x, y, a)$ tuples where $t$ denotes the timestamp, $x, y$ denote the 2D ground plane coordinate, and $a$ denotes the latest action or event label.
From $T$, we can compute the compliance rate or compare with the ground truth for evaluation metrics.

\section{Experiments} 

Our goal is to build a computer vision system for automatically assessing hand hygiene compliance across an entire hospital unit.
We conducted two primary experiments.
First, we evaluate our system's ability to assess hand hygiene compliance.
Our system's detections are compared against the ground truth, including a comparison to existing hand hygiene assessment solutions.
Second, we evaluate pur hand hygiene activity classifier.
Formulated as a binary classification task, we show quantitative and qualitative results.

\subsection{Baselines}
\textbf{Covert Observation.} Today, many hospitals measure hand hygiene compliance using \textit{secret shoppers} \citep{morgan2012automated}.
Secret shoppers are trained individuals who physically go to hospitals and measure compliance rates in-person without revealing the true purpose of their visit.
We refer to this as a \textit{covert observation}, as opposed to an \textit{overt observation} performed by someone openly disclosing their audit.
The purpose of covert observations is to minimize the Hawthorne effect \citep{adair1984hawthorne}.

In this work, we conducted two covert observational studies: (i) a single auditor responsible for monitoring the entire unit and (ii) a group of three auditors with a collective responsibility of the entire unit.
We refer to these groups as \textit{covert1} and \textit{covert3}, respectively.
The motivation for having two different covert groups is to measure the performance gains of having additional people for in-person observations.
Both covert groups were disguised as hospital visitors.
The group of three was spread out over the unit, remaining stationary, while the individual auditor constantly walked around the unit while monitoring hand hygiene compliance.

\textbf{Proximity Algorithm.}
Existing RFID solutions can be interpreted as proximity algorithms.
In some radio-based hand hygiene compliance assessments, a healthcare worker must \textit{badge in} before or after washing their hands upon entering or exiting a patient a room.
If a healthcare worker approaches a radio base station within some threshold, the RFID will activate indicating a localization event.
Since the standard error for radio-based positioning is one meter \citep{whitehouse2007practical, alahi2015rgbw, zanca2008experimental}, we use this as our activation threshold.
The proximity algorithm simulates this process.
If a person approaches within one meter of a patient's door or hand hygiene dispenser, they are considered compliant with hand hygiene protocol.
Using depth sensors, we can detect if a person approaches within one meter of a door or dispenser.

\subsection{Compliance Results}

\begin{figure*}[t]
\vspace{-8mm}
	\begin{minipage}{.47\linewidth}
		\centering
		\small
		\begin{tabular}{l|c} \toprule
			Method & Accuracy \\ \midrule
			Proximity Algorithm & 18\% \\
			Covert Observation (1 Person) & 63\% \\
			Covert Observation (3 People) & 72\% \\ \midrule
			\textbf{Our Model} & \textbf{75}\% \\
			\bottomrule
		\end{tabular}
		\caption{Comparison of hand hygiene assessment method. Each method must identify: (i) when a person enters or exit a room and (ii) classify the person as \textit{compliant} or \textit{non-compliant}. If the method correctly classifies a person within five seconds of the ground truth, the method scores a correct detection towards the accuracy metric.}\label{fig:accuracy}
	\end{minipage}%
	\hspace{3mm}
	\begin{minipage}{.51\linewidth}
		\centering
		\includegraphics[width=1.0\textwidth]{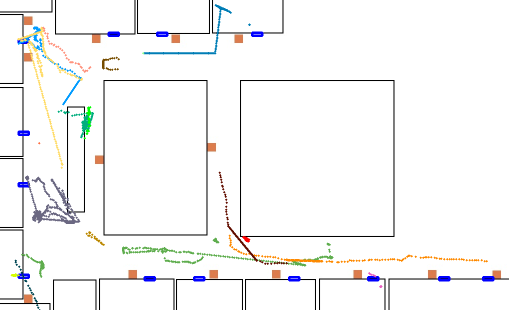}
		\caption{Top-down view of tracks. Blue rectangles are doors, orange squares are dispensers, and black lines are walls. Different track colors denote different people.}
		\label{fig:tracks}
	\end{minipage} 
\end{figure*}

We evaluated the result of our computer vision system compared to covert observational studies and the proximity algorithm in Figure \ref{fig:accuracy}.
Each compliance assessment method aims to identify when a person enters or exit a room.
Additionally, each compliance assessment method must classify the person as \textit{compliant} or \textit{non-compliant}.
The person is flagged as non-compliant if they do not obey proper handwashing protocol before entering or after leaving a patient room.
If the compliance assessment correctly identifies a compliant or non-compliant behavior, it is considered a successful detection and will increase its accuracy.
When compared to the ground truth, our model achieves a 75\% accuracy.
This is higher than both covert1 and covert3 observations. Although the covert3 observation achieves 72\% accuracy, in practice, this type of observation is rare.
Most covert observations planned by hospitals consists of a single covert observer.
In our experiments, a single covert observer achieved a 63\% accuracy.
This is because the single observer was simply unable to monitor the entire hospital unit by themselves.

\textbf{Traffic Patterns.} Using our model's pedestrian tracking ability, we can derive insights about traffic patterns in the physical space.
The method proposed in Section \ref{sec:method} allows us to convert between three-dimensional positions and two-dimensional coordinates for each person, at each point in time.
Figure \ref{fig:tracks} shows several tracks (i.e., sequence of 2D points denoting a person's location over time) from a top view perspective.
This type of analysis can be combined with hospital manifests to identify reasons for crowding around certain patient rooms and potentially identify areas prone to HAIs.
With such a visualization, we are able to intuitively understand the location distribution of healthcare workers.

\subsection{Hand Hygiene Activity Classification Results}

When trained from scratch, our experiments show that ResNet-152 outperforms the other model architectures, due to the high number of layers and parameters of the model.
We also show that augmenting the depth map with both the foreground segmentation and pose information also improves classifier performance.
By jointly training the classifier with a spatial transformer network, we see a $1\%$ boost in accuracy.

\begin{figure*}[t!]
	\centering
	\includegraphics[width=0.75\textwidth]{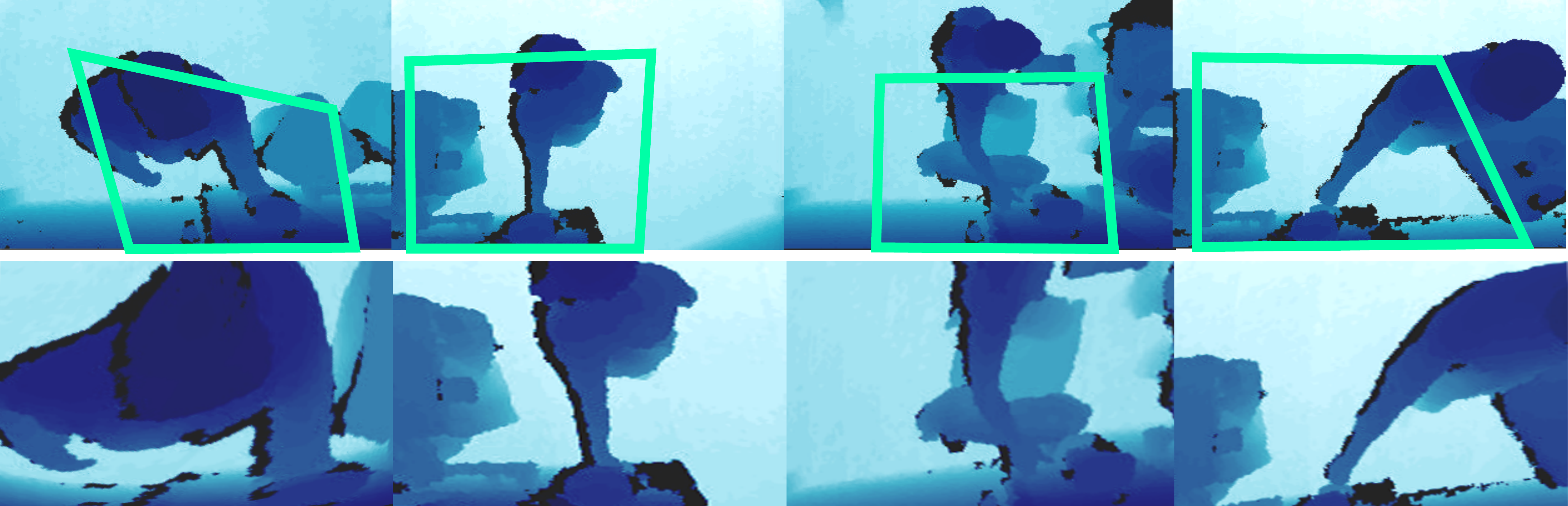}
	\caption{Examples before and after the spatial transformation. (Top) Input images with green bounding boxes from the grid generator. (Bottom) Transformed inputs. The model stretches, skews, and interpolates the input differently, depending on the scene contents.}
	\label{fig:stn_examples}
\end{figure*}

We now turn to a qualitative analysis of the hand hygiene classifier..
Figure \ref{fig:stn_examples} shows several before and after images produced by our model's transformation layer.
Because the transformation is explicitly parametrized (see Equation \ref{eq:affine}), we can recover the bounding box and overlay it on the original image.
Bounding boxes indicate the regions selected by our model, which are transformed and shown in Figure \ref{fig:stn_examples}.
Our model is capable of ``zooming in" on areas of visual importance.
The resulting transformation contains fewer extraneous and redundant information (i.e., floor and walls).
We want to emphasize that we provide no information about sensor or dispenser placement to our model.
The model learns how to transform each input image solely based on patterns present across and within viewpoints.
Such a visualization greatly assists the interpretability of our model to clinicians.

\begin{table}[t]
	\centering
	\small
	\vspace{-3mm}
	\begin{tabular}{l|ccc|c|cccc} \toprule
		Architecture & D & F & P & STN & Accuracy & Precision & Sensitivity & Specificity \\ \midrule
		AlexNet & \checkmark & \checkmark & \checkmark & \checkmark & 93.9 & 91.8 & 96.3 & 91.4 \\
		VGG-16 & \checkmark & \checkmark & \checkmark & \checkmark & 92.3 & 91.9 & 92.8 & 91.8 \\ 
		ResNet-152 & \checkmark & \checkmark & \checkmark & \checkmark & \textbf{95.5} & 94.6 & \textbf{96.7} & \textbf{94.5} \\ \midrule
		ResNet-152 & \checkmark & \checkmark &  &  & 94.6 & 93.1 & 96.3 & 92.9 \\
		ResNet-152 & \checkmark & \checkmark &  & \checkmark & 95.4 & \textbf{95.3} & 96.6 & 94.2 \\
		 \bottomrule
	\end{tabular}
	\caption{Classifier ablation study: Effect of different inputs and architectures. D, F, and P denote depth, foreground, and pose inputs, respectively. STN denotes the spatial transformer network. The training and test sets are balanced with a 50/50 class-split.}
	\label{table:results}
\end{table}

\section{Discussion \& Related Work}
There has been recent interest in creating smart hospitals with the aim of increasing operational efficiency and improving patient care \citep{ma2017measuring, twinanda2015data, fisher2008tracking, gao2006vital, noury2008level, wu2006ewellness, nugent2006system}.
One use case of smart hospitals is in the prevention of HAIs \citep{cook2009assessing}, or more specifically, for monitoring and tracking hand hygiene of hospital staff \citep{nishimura1999handwashing}.
Current technologies that track hand hygiene include RFID-based systems \citep{fuhrer2006building}.
However such systems are limited in resolution and precision \citep{alahi2015rgbw, zanca2008experimental}.
Computer vision-based tracking systems have shown promising results in non-clinical applications such as self-driving cars \citep{cho2014multi} and sports analytics \citep{halvorsen2013bagadus}.
Several works have applied computer vision to hospitals for several tasks and have shown promising results \citep{chakraborty2013video, ma2017measuring, twinanda2015data}.

In this paper, we proposed a non-intrusive and large-scale vision-based system for tracking people's activity in hospitals.
We evaluated our method on measuring hand hygiene compliance and showed hand hygiene activity classification results.
Our method outperforms existing solutions such as proximity-based techniques and covert in-person observational assessments.
We presented intuitive, qualitative results that analyze human movement patterns and conduct spatial analytics which convey our method's interpretability.
The system presented in this work is a step towards a vision-based smart hospitals and demonstrates promising results for reducing HAIs and ultimately improve the quality of patient care.

\newpage
\clearpage
\bibliography{main}

\end{document}